# Contrastive Siamese Representation Learning for Predictive Maintenance of Electrical Submersible Pumps

**Seshu K. Damarla*, Xiuli Zhu****

**Department of Chemical and Materials Engineering, University of Alberta, Edmonton, Alberta, Canada (damarla@ualberta.ca)*
***Department of Optical-Electrical and Computer Engineering, University of Shanghai for Science and Technology, Shanghai, P. R. China (xiulizhu@usst.edu.cn)*

**Abstract**: Electrical submersible pumps (ESPs) are essential in offshore oil production, where unexpected failures can result in significant operational and financial losses. Accurate predictive maintenance for ESP systems remains challenging due to nonlinear operating conditions, class imbalance, and variability among pump units. To address these issues, this study presents a fault diagnosis framework that incorporates class imbalance awareness by employing Siamese contrastive representation learning and prior-corrected k-nearest neighbor (KNN) classification. The method first extracts discriminative features relevant to fault detection from vibration-domain indicators and engineered harmonic relationships. A Siamese neural network is trained with class-balanced contrastive pairs to construct an embedding space that clusters samples of the same fault type and separates different fault classes. To further mitigate class imbalance during classification, a prior-corrected distance-weighted KNN is applied. The framework is validated using a Leave-One-ESP-Out (LOEO) strategy to evaluate generalization to previously unseen ESP units. Experimental results indicate that the proposed framework delivers robust and consistent fault classification performance under realistic industrial conditions, supporting its potential for reliable predictive maintenance and intelligent ESP system monitoring.

## 1. INTRODUCTION

Of late, the oil and gas industry has shifted significantly toward artificial lift as the primary production method. This shift is primarily due to the depletion of easily accessible reservoirs and the increasingly challenging conditions of oil wells (Brown (1982)). An electric submersible pump (ESP) is a critical artificial lifting equipment used to extract oil and gas from offshore platforms. ESPs are used across a range of well types, including deep, directional, and water-producing wells, due to their strong lifting capacity, substantial displacement, and high lift (Liang et al. (2015)). These wells are located 2,000 meters below the seafloor, making equipment maintenance impractical after installation. Failures in ESPs lead to significant financial losses due to high equipment replacement costs and prolonged interruptions in oil production. Prompt detection of irregularities in ESP operation is critical for maintaining the safety and reliability of the oil recovery process. Implementing effective monitoring strategies is necessary to mitigate risks and ensure continuous ESP operation in oil production (Takacs (2017)).

Laboratory-based vibration tests conducted prior to ESP deployment mitigate the risk of equipment failure. During these tests, multiple accelerometers are positioned at various locations on the ESP to collect vibration signals under varying operating conditions over extended periods (for instance, 3 days). Following the tests, specialists employ computational tools to visually assess the vibration spectra and determine whether the equipment operates within acceptable parameters. Detecting each vibratory problem at an early stage prevents failure, thereby reducing the need for subsea intervention and minimizing production losses and associated costs. While this procedure mitigates the risk of failure following system deployment, it is highly dependent on experts with extensive experience in interpreting vibration spectra. The specialized knowledge required is not easily transferable and demands years of practical training. Consequently, a shortage of such specialists can result in project delays and may cause less experienced technicians to
approve equipment that is not suitable for operation. To reduce this reliance on human expertise, previous studies have addressed the ESP fault diagnosis problem using various machine learning methods.

In Li et al. (2025), the authors presented a novel unsupervised anomaly detection method, DSTCL-ATG, for ESP systems that combines spatio-temporal contrastive learning with adaptive threshold generation. The proposed approach achieved more accurate and earlier fault detection with fewer false alarms than existing state-of-the-art methods on real oilfield ESP datasets. In Khalili et al. (2025), the authors proposed a comprehensive rule-based expert system for real-time fault diagnosis in ESP systems, comprising 500 interpretable diagnostic rules that integrate sensor data, statistical analysis, machine learning insights, and field expertise. The framework achieved 94.3% fault-detection accuracy while rapidly generating alerts, providing a transparent, scalable solution to improve ESP reliability and reduce unplanned downtime. A fibre-optic Raman Distributed Temperature Sensing (RDTS) method was introduced in Torelli et al. (2025) for monitoring temperature distribution in ESP motors under varying load conditions. Experimental results showed that RDTS enables continuous and accurate thermal monitoring, helping identify overheating regions, improve cooling analysis, and support early fault detection in harsh oilfield environments where conventional sensors are limited. Rauber et al. (2013) utilized support vector machines with frequency-domain statistical features for ESP fault diagnosis on the ESPset dataset, which included 1,844 samples across three classes: normal, misalignment, and unbalance. The results were compared to those of neural networks, with no significant performance difference observed. Boldt et al. (2014) introduced an extreme learning machine combined with feature selection for unbalance fault detection, using 5,314 samples and achieving superior performance relative to other machine learning methods on the same statistical features. In Boldt et al. (2015), a more efficient feature selection method was presented for ESP fault diagnosis, resulting in minimal accuracy loss on the same dataset as Boldt et al. (2014). Furthermore, Khalili et al. (2025) performed a comparative analysis of machine learning algorithms using eight expert-designed hand-crafted features on a five-class imbalanced ESP dataset with 4,570 samples.

All these works used proprietary ESP data, and the traditional k-fold cross-validation approach was employed to evaluate their methodologies. The traditional k-fold cross-validation is not appropriate for ESP fault diagnosis because samples from the same ESP unit may appear in both the training and test sets. Measurements from identical equipment typically exhibit similar operating characteristics and fault patterns, which increases the risk of data leakage and results in overly optimistic performance estimates. In the present work, we propose a novel and practical approach for fault diagnosis of ESPs. The main contributions of this work are provided below:

- A Siamese embedding network was developed and trained using contrastive loss to learn representations that discriminate between different fault types.
- A top-k-per-class distance-based classification strategy was introduced to mitigate the dominance of majority classes during inference.
- A leave-one-ESP-out (LOEO) validation protocol was implemented to assess cross-ESP generalization on previously unseen pumps.
- Dynamic macro F1 evaluation was adopted to address fragmented industrial datasets in which not all fault classes are present in each test ESP.
- The proposed approach demonstrated enhanced robustness for ESP fault diagnosis under highly imbalanced and domain-shifted industrial operating conditions.

The rest of the manuscript is organized as follows. In Section II, ESP operation and data collection are discussed. Section III provides a detailed discussion of the proposed methodology. Section IV presents results and analysis. Finally, the paper is concluded in Section V.

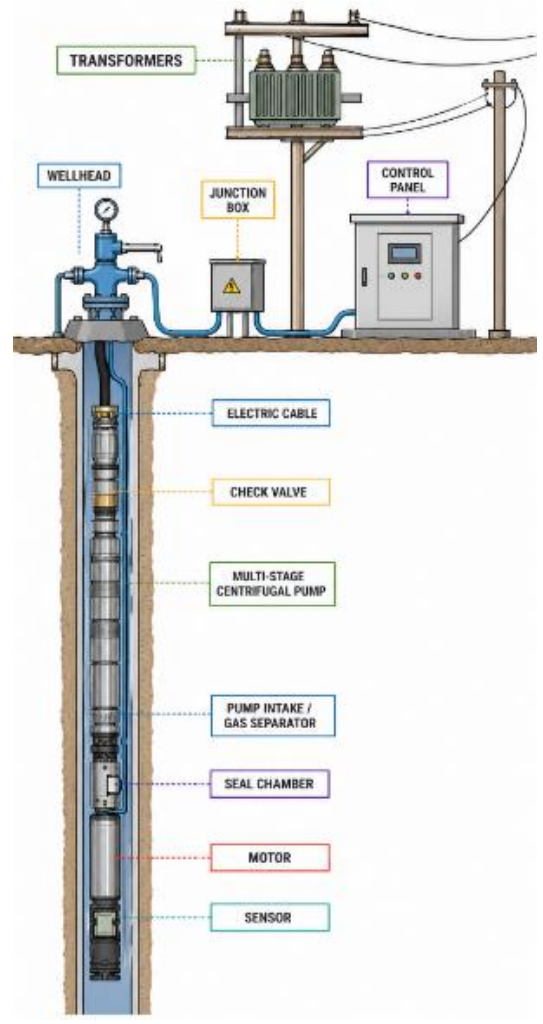

Figure 1. Electrical submersible pump.

## 2. ESP OPERATION AND DATA COLLECTION

The ESP system illustrated in Figure 1 consists of multiple centrifugal pump stages connected to a submersible electric motor. Heavy-duty cables link the motor to surface control systems to provide power. The motor rotates a shaft attached to the pump, and as the impellers turn, they draw fluid through the pump intake, pressurize it, and transport it to the surface. ESPs are commonly employed in low-flow oil wells, however, their installation, operation, and maintenance incur substantial costs. Consequently, comprehensive testing and condition monitoring are conducted prior to deployment to mitigate the risk of unexpected failures and related economic losses.

The present work uses publicly available ESP dataset (Varejão et al. (2024)), which was obtained by conducting experiments in specially designed testing wells in a laboratory. During testing, vibration sensors were installed on different sections of the ESP assembly, including the motor, pump, and protector, to capture vibration behavior under various operating conditions. The ESPs were operated at different motor speeds, flow conditions, and electrical supply frequencies, while the acquired vibration signals were transformed into the frequency domain and analyzed by experts to identify characteristic patterns associated with normal and faulty operating conditions. The ESP dataset contains vibration signals collected from 11 ESP systems operating under five conditions: normal operation, faulty sensor, misalignment, rubbing, and unbalance. Each vibration spectrum was labeled by a human expert through visual inspection of the spectral patterns. The complete dataset contains 6032 vibration signal instances and exhibits significant class imbalance, which is common in industrial fault diagnosis applications. Due to the confidential nature of the industrial vibration data, only processed and normalized spectral information is publicly released, rather than the original raw signals. Table I presents the number of vibration spectra available for each fault category across the 11 ESPs. Figure 2 illustrates representative vibration spectra corresponding to the five operating conditions obtained from ESP 1.

Table I. Class Distribution of the ESP Dataset

| ESP | NL | FS | MA | RB | UB |
|---|---|---|---|---|---|
| 1 | 201 | 0 | 0 | 0 | 51 |
| 2 | 644 | 23 | 0 | 46 | 34 |
| 3 | 277 | 10 | 0 | 12 | 1 |
| 4 | 347 | 11 | 0 | 56 | 6 |
| 5 | 578 | 60 | 52 | 7 | 103 |
| 6 | 614 | 79 | 0 | 79 | 58 |
| 7 | 272 | 4 | 0 | 0 | 102 |
| 8 | 612 | 0 | 0 | 0 | 72 |
| 9 | 309 | 108 | 0 | 0 | 69 |
| 10 | 308 | 0 | 18 | 22 | 73 |
| 11 | 639 | 0 | 0 | 68 | 7 |
| Overall | 4801 | 295 | 70 | 290 | 576 |

NL-normal operation, FS-faulty sensor, MA-misalignment, RB-rubbing, UB-unbalance

## 3. THE PROPOSED METHODOLOGY

Figure 3 depicts the configuration of the proposed methodology. The methodology comprises representation learning and inference via the modified k-nearest neighbors. A Siamese neural network is employed to learn representations (embeddings) of the vibration data or spectra. Because the ESP data is highly imbalanced, the network is trained on pairs of training samples rather than on individual samples. Positive and negative sample pairs are created. Two samples from the same class form a positive pair:

$$(x_i, x_j), \qquad y_{ij} = 1$$

Two samples from different classes generate a negative pair:

$$(x_i, x_j), \qquad y_{ij} = 0$$

Table 1 shows that the normal-condition samples outnumber those of the fault classes, which may bias representation learning towards the majority class. To alleviate this problem, class-balanced pair sampling is employed. For each class, an equal number of positive and negative pairs is generated. Therefore, the minority fault classes contribute uniformly during training.

The network consists of two branches sharing the same encoder. The encoder is constructed using fully connected layers and dropout layers. The first branch receives its input $x_i$ and the input is processed in the first fully connected layer. Neurons in the fully connected layers use ReLU activation functions, which aid the encoder learn nonlinear fault-discriminating embeddings from the spectral data. The ReLU activation function is preferred over other functions because it is computationally simple, trains faster and reduces vanishing gradient problems. The output of the first fully connected layer is fed to the dropout layer, which helps prevent overfitting in neural networks. During training, the dropout randomly deactivates a fraction of the neurons in the fully connected layer; i.e., the weights of deactivated neurons are not updated. The output of the first dropout layer is passed on to the subsequent layers of the encoder. The last layer of the encoder is the embedding layer that learns representations/embeddings ($z_i$) from the input. In the same way, the second branch of the encoder learns embeddings $z_j$ from its input $x_j$. Both $z_i$ and $z_j$ are normalized to ensure they are on the same scale for stabilizing network training, permitting scale-independent distance comparisons, and improving KNN classification. The Euclidean distance between $z_i$ and $z_j$ is computed:

$$d_{ij} = \|z_i - z_j\|_2 \qquad (1)$$

The trainable weights of the Siamese network are learned by minimizing the contrastive loss:

$$L = y_{ij} d_{ij}^2 + (1 - y_{ij}) max(0, m - d_{ij})^2 \qquad (2)$$

where $y_{ij} = 1$ for positive pairs, $y_{ij} = 0$ for negative pairs, and $m$ is the margin hyperparameter.

The contrastive loss function performs two actions concurrently. If the input feature vectors $x_i$ and $x_j$ belong to the same class, then $y_{ij}$ becomes 1, and the loss function tries to minimize $d_{ij}$ so that the corresponding embeddings are pulled closer. If $x_i$ and $x_j$ are from different classes, then $y_{ij}$ is set to 0 to push the respective embeddings apart. As the training progresses, samples from the same class are grouped together while samples from different classes are separated. Once the Siamese network is trained, it serves as a feature extractor. All training and testing samples are mapped into the learned embedding space to obtain embedding or representation vectors.

$$Z_{train} = encoder(X_{train}) \qquad (3)$$

$$Z_{test} = encoder(X_{test}) \qquad (4)$$

These embeddings contain fault-discriminative representations learned by optimizing the contrastive loss function in (2).

To classify the embeddings, we propose a modified KNN: top-k-per-class KNN. A test sample $x_q$ is fed to the encoder and an embedding $z_q$ is obtained and then normalized. Distance between $\tilde{z}_q$ and each training embedding is computed. For each class, the $k$ nearest samples belonging only to that class are selected. Mean class-wise distance is computed as shown in the following equation.

$$score(c) = \sum_{i=1}^{k} d_i \qquad (5)$$

where $d_i$'s are the nearest distances from class $c$.

The class predicted for $\tilde{z}_q$ is given by

$$\tilde{y} = arg \min_{c} score(c) \qquad (6)$$

The modified KNN reduces the dominance of the majority class by assigning equal weight to each class in decision-making. .

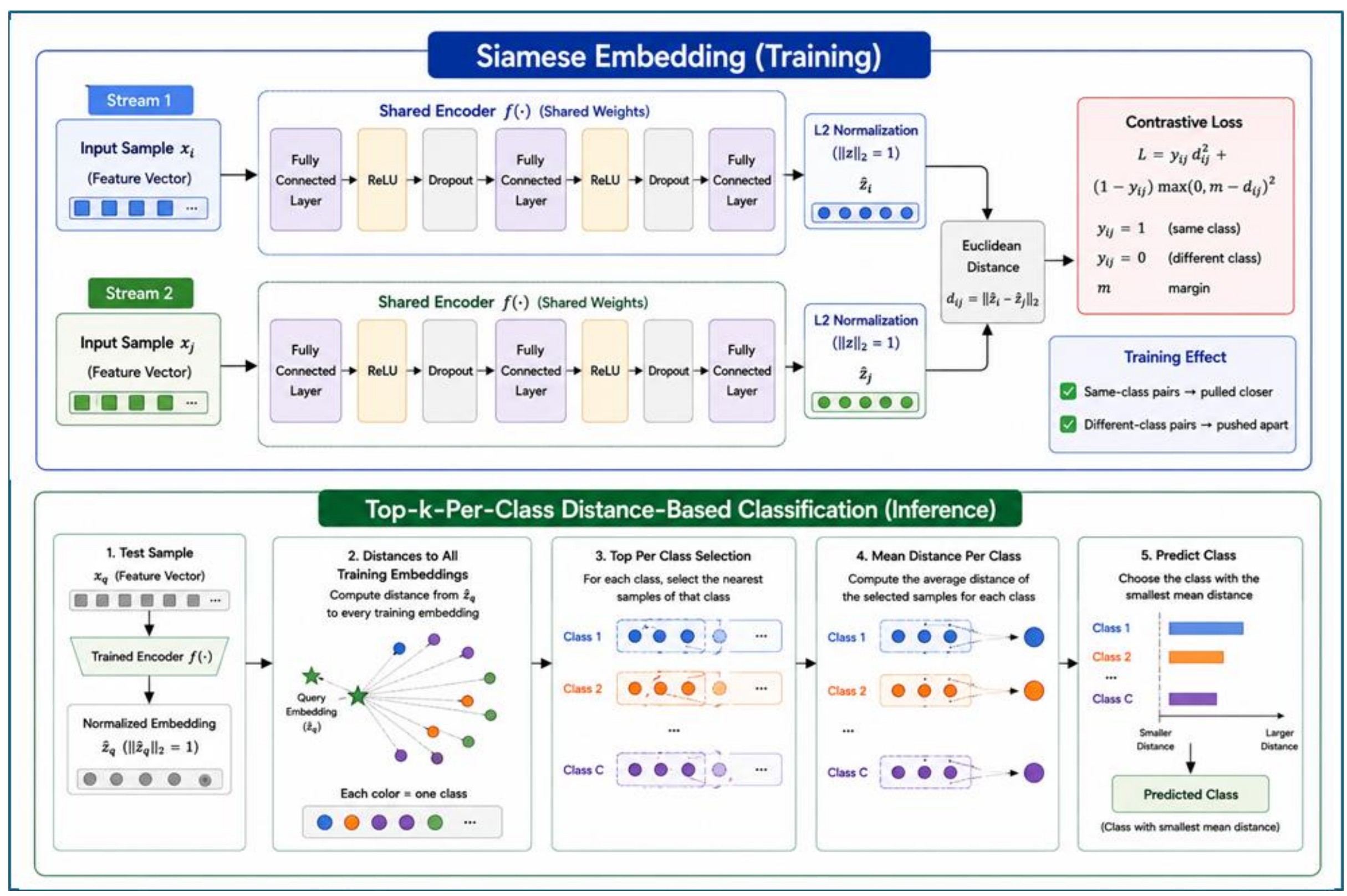

Figure 3. Configuration of proposed methdology.

## 4. RESULTS AND DISCUSSIONS

Each vibration spectrum contains 12150 data points, and processing the raw spectra requires a complex Siamese network, making it computationally expensive. Handcrafted vibration features were extracted from specific frequency regions of the vibration spectra. The features: $median(8,13)$ and $median(98,102)$ represent the median spectral amplitudes within the frequency intervals $(8\%X, 13\%X)$ and $(98\%X, 102\%X)$, respectively, where $X$ denotes the rotating frequency. The feature: $rms(98,102)$ is defined as the root mean square value of the amplitudes within the $(98\%X, 102\%X)$ interval. Additionally, $peak1x$ and $peak2x$ correspond to the spectral amplitudes at the fundamental rotating frequency $1X$ and its second harmonic $2X$, respectively.

The coefficients $a$ and $b$ are derived from an exponential regression model of the form:

$$e^{(aA+b)} \qquad (7)$$

Here, A represents an array of uniformly spaced relative frequencies up to $0.4X$, excluding zero. These features provide compact, fault-sensitive representations of the vibration data and were derived according to the feature extraction methodology described in Varejão et al. (2024). After handcrafted feature extraction, the vibration database size became 6032×9, where the last two dimensions represent class label and ESP index, respectively. As explained in Section I, the traditional k-fold cross-validation is not suitable for the current ESP database because vibration spectra collected from the same ESP may appear in both training and test sets. Therefore, we propose a leave-one-ESP-out (LOEO) cross-validation approach. According to this approach, in each fold, one ESP is kept for testing, while the remaining ESPs are used for training. For instance, in the first fold, ESP 1 acts as the test ESP and ESPs 2 through 11 are designated for training. This process is repeated until each ESP has been used once as the test ESP, as summarized in Table II. Within each fold, the robust scaler provided in (8) was applied to the handcrafted features.

$$X_{scaled} = \frac{X - meadian(X)}{IQR(X)} \qquad (8)$$

where IQR is the interquartile range.

The robust scaler was used because the vibration signals may contain outliers and exhibit non-Gaussian distributions. In each fold, the robust scaler was fitted only to the training ESPs, and features from a test ESP were scaled using the fitted scaler. This approach prevented information leakage. The configuration and training details of the proposed methodology are provided in Table III. Figure 4 shows the performance of the Siamese encoder during training within each fold. Figure 5 compares the t-SNE visualization of the handcrafted feature space with that of the learned embedding space for the training data in Fold 1. In the handcrafted feature space, substantial class overlap results in poorly defined boundaries. This observation suggests that handcrafted features alone do not provide sufficiently discriminative information for reliable fault classification. In contrast, the learned embedding space exhibits improved class-wise separation. Samples from the same class are more densely clustered, while the samples from different classes are more distinctly separated. This outcome indicates that the Siamese network effectively learned fault-discriminative features by minimizing intra-class distances and maximizing inter-class distances via optimization of the contrastive loss function.

Table II. Training and testing ESPs under LOEO cross-validation

| Fold | Training ESPs | Test ESP |
|---|---|---|
| 1 | 2-11 | 1 |
| 2 | 1, 3-11 | 2 |
| 3 | 1, 2, 4-11 | 3 |
| 4 | 1, 2, 3, 5-11 | 4 |
| 5 | 1-4, 6-11 | 5 |
| 6 | 1-5, 7-11 | 6 |
| 7 | 1-6, 8-11 | 7 |
| 8 | 1-7, 9-11 | 8 |
| 9 | 1-8, 10, 11 | 9 |
| 10 | 1-9, 11 | 10 |
| 11 | 1-10 | 11 |

Table III. Configuration and Training Details

| Component | Configuration / Training Details |
|---|---|
| Number of FC layers | 3 |
| No. of neurons in each FC layer | 64 |
| No. of neurons in the embedding layer | 16 |
| No. of ReLU layers | 2 |
| No. of dropout layers | 2 |
| Dropout rate | 0.15 |
| Embedding normalization | L2 normalization |
| Loss function | Contrastive loss |
| Margin parameter | 1.0 |
| Pairs generated per class | 500 |
| Training optimizer | Adam |
| Learning rate | 1 × 10^-3 |
| Weight decay | 1 × 10^-4 |
| Batch size | 128 |
| Number of training epochs | 200 |
| Feature scaling | RobustScaler |
| No. of nearest neighbors per class | 3 |
| Python frameworks | PyTorch, Scikit-learn |

As shown in Table I, some ESPs do not contain all classes. As a result, conventional macro metrics penalize the methodology's performance for missing classes that do not exist in the current test ESP. Therefore, dynamic evaluation metrics were adopted in the present work:

- For each fold, determine the unique classes present ($C_{present}$) in the ground truth labels of the test ESP.
- Metrics: precision, recall and F1 score are calculated only for the classes in $C_{present}$.
- Macro average is computed using only the number of present classes rather than the number of theoretical classes.

$$Dynamic\ macro\ F1 = \frac{\sum_{c \in C_{present}} F1_c}{C_{present}} \quad (9)$$

Table IV provides the dynamic evaluation metrics computed in each fold of the LOEO cross-validation. The proposed methodology achieved a mean accuracy of 0.9100, with mean recall, precision, and F1-score values of 0.7776, 0.8208, and 0.7754, respectively. These findings demonstrate that the framework effectively learns discriminative fault representations that generalize to previously unseen ESPs.

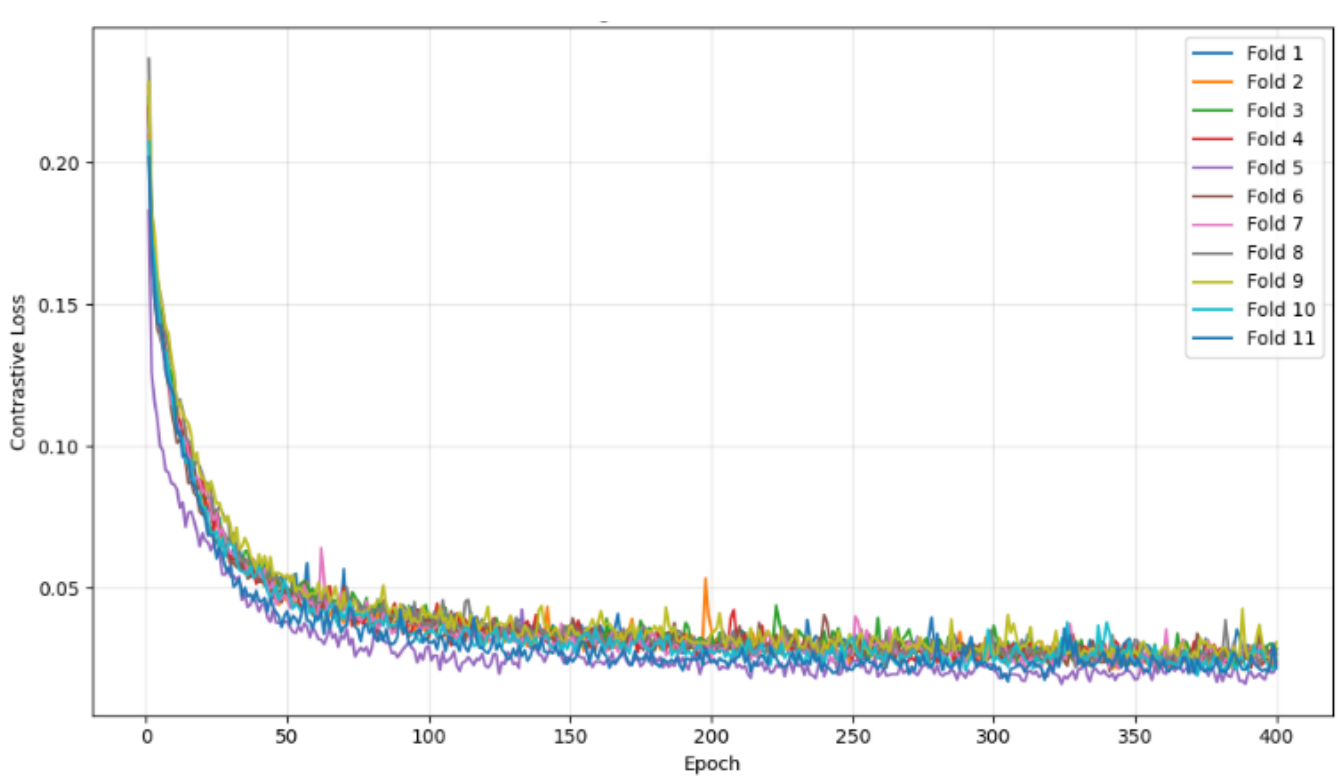

Figure 4. Learning curves across LOEO folds.

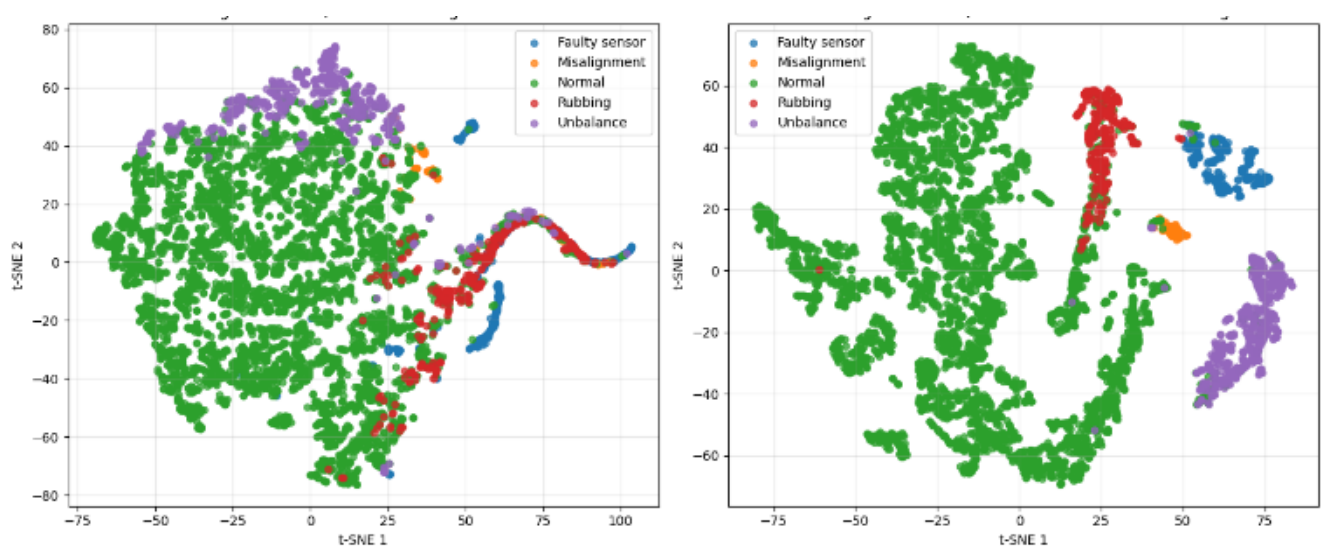

Figure 5. t-SNE visualization of original handcrafted features and learned Siamese embeddings for training data in Fold 1.

Table IV. Fold-wise classification performance

| Fold | Accuracy | Recall | Precision | F1 score |
|---|---|---|---|---|
| 1 | 0.956349 | 0.914106 | 0.980583 | 0.943933 |
| 2 | 0.947791 | 0.810239 | 0.858927 | 0.831417 |
| 3 | 0.990000 | 0.937500 | 0.873208 | 0.880053 |
| 4 | 0.938095 | 0.800557 | 0.939867 | 0.844562 |
| 5 | 0.802500 | 0.646052 | 0.752272 | 0.564565 |
| 6 | 0.949398 | 0.870778 | 0.952554 | 0.904209 |
| 7 | 0.925926 | 0.793301 | 0.762600 | 0.775515 |
| 8 | 0.982456 | 0.971814 | 0.939832 | 0.955074 |
| 9 | 0.983539 | 0.987618 | 0.980381 | 0.983822 |
| 10 | 0.850356 | 0.562520 | 0.566544 | 0.557660 |
| 11 | 0.683473 | 0.258945 | 0.421500 | 0.288793 |
| mean | 0.909989 | 0.777584 | 0.820752 | 0.775418 |

## 5. CONCLUSIONS

In the present work, we introduced a robust fault-diagnosis framework for electrical submersible pumps (ESPs) by leveraging Siamese contrastive representation learning and a modified top-k-per-class k-nearest neighbors classifier. This approach addressed key industrial challenges: class imbalance, variability among ESP units, and the requirement to generalize effectively to previously unseen pumps. Training the Siamese network with class-balanced contrastive pairs enabled the model to learn more discriminative feature representations, thereby improving separation between the fault categories.

The method was evaluated under realistic industrial conditions using a leave-one-ESP-out (LOEO) validation strategy, which prevented data leakage between the training and testing ESPs, unlike the conventional k-fold cross-validation. The dynamic macro evaluation metrics were also employed to address the fragmented distribution of the fault classes across the ESP units. The experimental results demonstrated strong classification performance, with mean accuracy, recall, precision, and F1-score values of 0.9100, 0.7776, 0.8208, and 0.7754, respectively. The proposed framework demonstrated enhanced robustness and improved generalization across different ESPs under realistic operating conditions, indicating strong potential for intelligent predictive maintenance and automated vibration-based fault diagnosis in offshore ESP systems.

## REFERENCES

Brown, K.E. (1982). Overview of artificial lift systems. *Journal of Petroleum Technology*, 34(10), 2384–2396.

de Assis Boldt, F., Rauber, T.W., Varejão, F.M., and Ribeiro, M.P. (2014). Performance analysis of extreme learning machine for automatic diagnosis of electrical submersible pump conditions. In *2014 12th IEEE International Conference on Industrial Informatics (INDIN)*, 67–72.

de Assis Boldt, F., Rauber, T.W., Varejão, F.M., and Ribeiro, M.P. (2015). Fast feature selection using hybrid ranking and wrapper approach for automatic fault diagnosis of motorpumps based on vibration signals. In *2015 IEEE 13th International Conference on Industrial Informatics (INDIN)*, 127–132.

Khalili, Y., Ahmadi, M., and Moraveji, M.K. (2025). A rule-based expert system for real-time fault diagnosis in electrical submersible pump systems. *Research Square*.

Khalili, Y., Ahmadi, M., and Moraveji, M.K. (2025). A comprehensive review of failure modes in electrical submersible pumps: Diagnosis, predictive maintenance, and engineer's guide. *Arabian Journal for Science and Engineering*, 50, 20445–20466.

Liang, X., He, J., and Du, L. (2015). Electrical submersible pump system grounding: Current practice and future trend. *IEEE Transactions on Industry Applications*, 51(6), 5030–5037.

Li, K., Li, S., Li, Q., Jiao, Z., Fu, J., Gao, X., and Zhang, L. (2025). Dual spatio-temporal contrastive learning network with adaptive threshold generation for anomaly detection of electric submersible pump. *IEEE Transactions on Instrumentation and Measurement*, 74, Art. no. 3548914.

Rauber, T.W., Varejão, F.M., Fabris, F., Rodrigues, A., and Ribeiro, M.P. (2013). Automatic diagnosis of submersible motor pump conditions in offshore oil exploration. In *IECON 2013 - 39th Annual Conference of the IEEE Industrial Electronics Society*, 5537–5542.

Takacs, G. (2017). *Electrical Submersible Pumps Manual: Design, Operations, and Maintenance*. Gulf Professional Publishing, Houston.

Torelli, G. de A., Cubas, J.M.C., de Pelegrin, J., Piazentin, R., Romero, G.A., Bacellar, F.R.R., da Silva, L.C.T., da Silva, J.A.G., Stel, H., Dreyer, U.J., Morales, R.E.M., and Cardozo da Silva, J.C. (2025). Experimental study on the thermal response of electric submersible pump motors to load variation using Raman distributed temperature sensing. *IEEE Sensors Letters*, 9(10), Art. no. 2504304.

Varejão, F.M., Mello, L.H.S., Ribeiro, M.P., Oliveira-Santos, T., and Rodrigues, A.L. (2024). An open source experimental framework and public dataset for vibration-based fault diagnosis of electrical submersible pumps used on offshore oil exploration. *Knowledge-Based Systems*, 288, 111452.